\documentclass[journal]{IEEEtran}
\usepackage{amsmath,amsfonts}
\usepackage{algorithmic}
\usepackage{algorithm}
\usepackage{array}
\usepackage[caption=false,font=normalsize,labelfont=sf,textfont=sf]{subfig}
\usepackage{textcomp}
\usepackage{stfloats}
\usepackage{url}
\usepackage{verbatim}
\usepackage{graphicx}
\usepackage{svg}
\usepackage{enumitem}
\usepackage{soul}

\usepackage[style=ieee,backend=biber]{biblatex}

\usepackage{amssymb} 
\usepackage[normalem]{ulem}
\usepackage{booktabs}
\usepackage{hyperref} 
\usepackage{cleveref}
\usepackage{multirow}
\usepackage{makecell}
\usepackage{bm}
\usepackage{xcolor}
\begin{document}

\title{Probabilistic Hysteresis Factor Prediction
for Electric Vehicle Batteries with Graphite Anodes Containing Silicon}

\author{%
Runyao Yu,
Viviana Kleine,
Philipp Gromotka,
Thomas Rudolf,
Adrian Eisenmann,
Gautham Ram Chandra Mouli,~\IEEEmembership{Member,~IEEE},
Peter Palensky,~\IEEEmembership{Senior Member,~IEEE},
and Jochen L.~Cremer,~\IEEEmembership{Member,~IEEE}%
\thanks{Runyao Yu and Jochen L.~Cremer are with Delft University of Technology, Delft, The Netherlands, and also with the AIT Austrian Institute of Technology, Vienna, Austria (e-mail: runyao.yu@tudelft.nl; j.l.cremer@tudelft.nl). Viviana Kleine is with Delft University of Technology, Delft, The Netherlands, and also with Porsche Engineering Group, Weissach, Germany (e-mail: v.kleine@student.tudelft.nl). Philipp Gromotka, Thomas Rudolf, and Adrian Eisenmann are with Porsche Engineering Group, Weissach, Germany (e-mail: philipp.gromotka@porsche-engineering.de; thomas.rudolf@porsche-engineering.de; adrian.eisenmann@porsche-engineering.de). Gautham Ram Chandra Mouli and Peter Palensky are with Delft University of Technology, Delft, The Netherlands (e-mail: g.r.chandramouli@tudelft.nl; p.palensky@tudelft.nl). Corresponding author: Runyao Yu.}%
}

\markboth{IEEE TRANSACTIONS ON TRANSPORTATION ELECTRIFICATION}
{}

\maketitle

\begin{abstract}
Batteries with silicon-graphite-based anodes, which offer higher energy density and improved charging performance, introduce pronounced voltage hysteresis, making state-of-charge (SoC) estimation particularly challenging. 
Existing approaches to modeling hysteresis rely on exhaustive high-fidelity tests or focus on conventional graphite-based lithium-ion batteries, without considering uncertainty quantification or computational constraints.
This work introduces a data-driven approach for probabilistic hysteresis factor prediction, with a particular emphasis on applications involving  silicon-graphite anode-based batteries. A data harmonization framework is proposed to standardize heterogeneous driving cycles across varying operating conditions. Statistical learning and deep learning models are applied to assess performance in predicting the hysteresis factor with uncertainties while considering computational efficiency. Extensive experiments are conducted to evaluate the generalizability of the optimal model configuration in unseen vehicle models through retraining, zero-shot prediction, fine-tuning, and joint training. By addressing key challenges in SoC estimation, this research facilitates the adoption of advanced battery technologies. A summary page is available at: {\color{orange}\url{https://runyao-yu.github.io/Porsche_Hysteresis_Factor_Prediction/}}
\end{abstract}


\section{Introduction}
\IEEEPARstart{B}{atteries} are one of the key components of EVs. Central to battery management is the estimation of SoC, which indicates the remaining usable battery energy. A more precise SoC estimation reduces the required SoC buffer, which is typically set to account for estimation uncertainties and protect the battery from aging. As a result, the SoC operating range can be expanded, enabling a greater depth of discharge and increasing the usable capacity. Especially, original equipment manufacturers (OEMs) continuously seek to improve the driving range. If the additional 2\% SoC is utilized to extend the range of vehicles, assuming an energy consumption of 19.8~kWh/100~km (comparable to an electric compact SUV), the range will be increased by approximately 10~km. Given that range anxiety remains a major barrier to EV adoption~\cite{petersen2022data, petersen2022fully, Samant2022EYStudy}, even small improvements can contribute to broader consumer acceptance. 
In addition to estimation accuracy, BMS is often memory- and computation-constrained in highly competitive cost structures as in the automotive industry, requiring efficient models suitable for limited Random Access Memory (RAM) and Read-Only Memory (ROM) capacities.

\begin{figure}[t]
    \centering
    \includegraphics[width=0.5\textwidth]{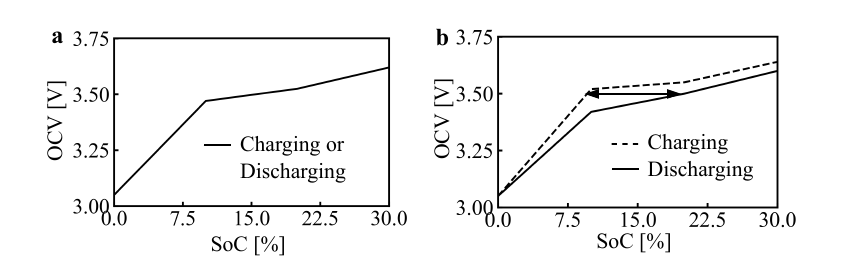}
    \caption{Illustration of OCV-SoC curves. 
    \textbf{a} The OCV-SoC relationship provided by the cell manufacturer.
    \textbf{b} The OCV-SoC relationship with voltage hysteresis, which is particularly significant for silicon–graphite anode-based batteries. }
    \label{fig:ocv_soc}
\end{figure}

Traditionally, cell manufacturers provide the Open-Circuit Voltage (OCV)-SoC reference table to correct the SoC value after battery relaxation, as illustrated in Fig. \ref{fig:ocv_soc} \textbf{a}, where a unique OCV corresponds to a unique SoC. However, OCV-SoC reference tables often depend on battery chemistry, as different electrode materials exhibit distinct voltage characteristics. Common materials in automotive applications are graphite for the anode \cite{Zhang2021GraphiteAdvances}, and LFP, NMC, and LMO for the cathode \cite{Ding2019AutomotivePerspectives}. Due to the limited energy density of graphite anodes, silicon has gained significant popularity as an alternative to graphite anodes \cite{InternationalEnergyAgencyIEA2023GlobalAmbitions}. Numerous automobile manufacturers announce plans to integrate silicon technology into their batteries, as integrating silicon as an anode enables longer vehicle ranges and faster charging times \cite{Evers2022WhyBatteries}. However, silicon-graphite anode-based batteries have certain drawbacks, with pronounced voltage hysteresis being a major limitation.

Voltage hysteresis represents a phenomenon where the OCV-SoC behavior differs based on the charge direction, as shown in Fig. \ref{fig:ocv_soc} \textbf{b}. During hysteresis, the OCV curve for charging consistently lies above that for discharging, making it challenging to determine the exact SoC from a given OCV value. For example, at an OCV of 3.5\,V, the SoC could correspond to either 10\% or 20\%. A hysteresis factor is introduced to address the issue, providing a quantifiable approach to SoC estimation within the hysteresis region, and is represented by a scalar continuous weighting towards the charging or discharging characteristic curve.
The non-linear and history-dependent nature of hysteresis further increases the computational complexity of SoC estimation models.

\begin{figure*}[t]
\begin{center}
\centerline{\includegraphics[width=1.0\linewidth]{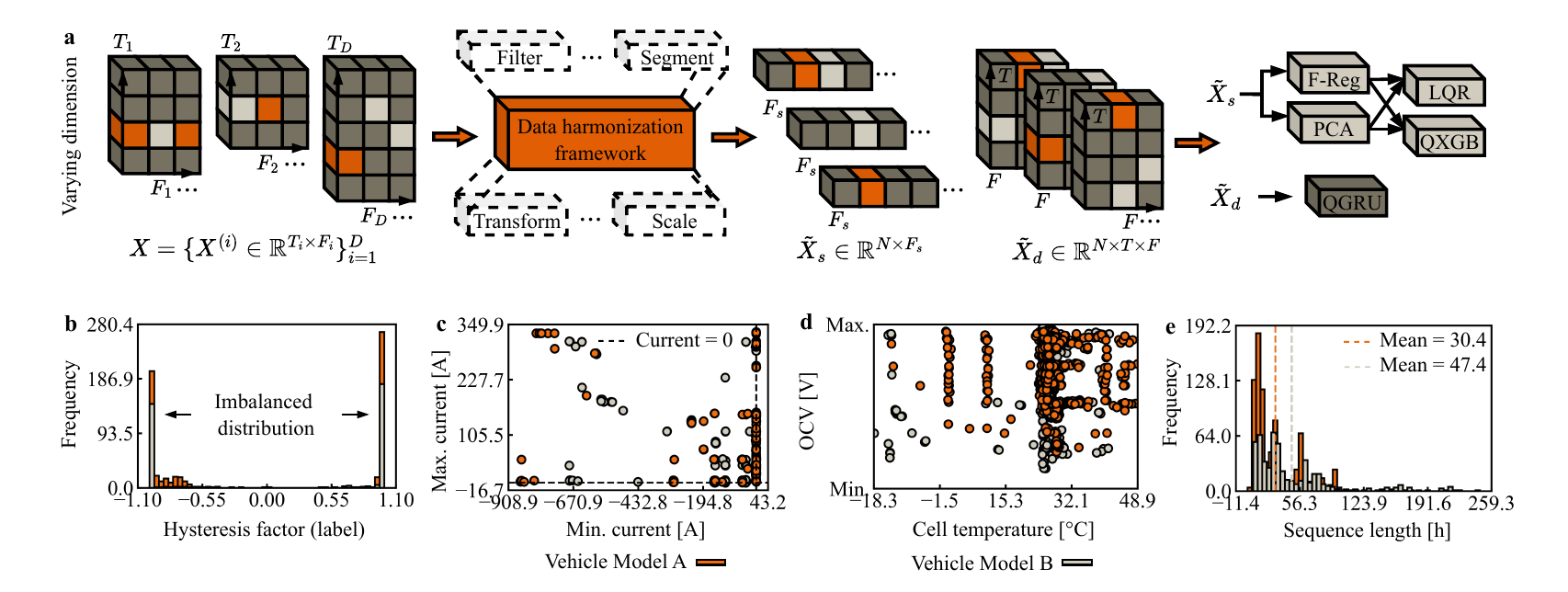}}
\caption{Overview of harmonization framework and exemplary driving cycles of two vehicle models. 
\textbf{a} Visualization of the harmonization framework. Heterogeneous driving cycle data 
are harmonized 
to produce unified representations: \( \tilde{X}_s \) and \( \tilde{X}_d \), which are fed into corresponding machine learning models. 
\textbf{b} Distribution of the hysteresis factor (range in [-1, 1]), showing a highly imbalanced distribution for vehicle mode A and vehicle model B. Less than 70\% of the data available for vehicle model~A is present in vehicle model~B.
\textbf{c} Scatter plot of maximum versus minimum current. The dashed lines indicate the zero-current baseline, separating charging (negative current) and discharging (positive current) regions, 
 where the vehicle model A exhibit a more aggressive charging and discharging behavior.
\textbf{d} Scatter plot of normalized OCV against cell temperature, capturing diverse thermal conditions and spanning a wide range from –18.3~°C to 48.9~°C. 
Most samples are concentrated between 20–35°C, suggesting that typical operations occur within standard ambient or thermally managed conditions.
\textbf{e} Distribution of sequence lengths of driving cycles, exhibiting a heterogeneous temporal structure.
The vehicle model A tends to be shorter on average, with a mean sequence length of 30.4~h, while the vehicle model B shows longer sequences, averaging 47.4~h. This suggests that different vehicle models are associated with different operating regimes.
}
\label{hdc}
\end{center}
\end{figure*}

In prior work, various models have been developed to incorporate hysteresis voltage into SoC estimation, which can be categorized into three main approaches: ECMs, physics-based models, and machine learning-based models. 
ECMs \cite{Plett2004ExtendedEstimation, Barai2015ACells, Rolt2019FullApplications, Xie2023ImprovedEffect, Xie2023State-of-ChargeModel, Xie2016EstimatingCharacteristic, Kwak2020ParameterEffect, Ko2021AStudy, Baronti2013ExperimentalBatteries, Dong2016OnlineMethod, Antony2023ABatteries, Kim2012OCVCell, Roscher2011DynamicBatteries, Zhou2022SOCBattery, Yu2022StudyModeling} incorporate hysteresis through various methods, such as defining a hysteresis factor to blend charge and discharge OCV curves or introducing additional parameters into the (dis)charging equations to account for hysteresis effects. 
Physics-based models \cite{Gao2023DeterminationBatteries} focus on modeling the anode’s open-circuit potential and isolating the hysteresis factor based on delithiation and lithiation processes. 
However, these models typically require special tests, such as constant current pulse tests, which are time-consuming and expensive \cite{Xu2020ImprovingTechnique}. 
To overcome these limitations, machine learning models \cite{Li2023ExploringBatteries, Xu2020ImprovingTechnique} have been developed, leveraging data-driven techniques such as random forests and Long Short-Term Memory (LSTM) networks to predict the hysteresis factor. 
Each of these modeling approaches aims to improve SoC estimation accuracy under hysteresis, with varying levels of model complexity and data requirements.

However, prior studies face several limitations that hinder their practical deployment and generalizability. 
First, data processing pipelines are often vaguely described and inconsistent across works \cite{Plett2004ExtendedEstimation, Barai2015ACells, Rolt2019FullApplications, Xie2023ImprovedEffect, Xie2023State-of-ChargeModel, Xie2016EstimatingCharacteristic, Kwak2020ParameterEffect, Ko2021AStudy, Baronti2013ExperimentalBatteries, Dong2016OnlineMethod, Antony2023ABatteries, Kim2012OCVCell, Roscher2011DynamicBatteries, Zhou2022SOCBattery, Yu2022StudyModeling, Gao2023DeterminationBatteries, Li2023ExploringBatteries, Xu2020ImprovingTechnique}, lacking a unified framework to systematically harmonize heterogeneous driving cycles under varying driving conditions. 
Second, most studies focus exclusively on graphite anodes, with only \cite{Gao2023DeterminationBatteries} explicitly addressing silicon-graphite chemistries, limiting the applicability to emerging battery technologies.
Third, existing approaches often produce point estimates without the quantification of predictive uncertainties using machine learning models \cite{Li2023ExploringBatteries, Xu2020ImprovingTechnique}, which can inherently adapt to non-stationary behaviors observed in real-world conditions. Without probabilistic modeling, however, it remains challenging to quantify prediction confidence under such dynamic scenarios.
Fourth, hardware constraints are typically overlooked. Without accounting for computational limitations, model implementations can lead to unrealistic computational demands, making them impractical for deployment in BMS with restricted processing power. 
Fifth, the generalization of hysteresis prediction models to unseen vehicle models remains largely unexplored. Given the variability in vehicle configurations, energy consumption profiles, and cell chemistries, assessing cross-vehicle transferability is essential.

To address the challenges of SoC estimation in the presence of voltage hysteresis—particularly in silicon-graphite anode-based batteries—we present a probabilistic machine learning approach built on a data harmonization framework. Our method explicitly accounts for predictive uncertainty and computational constraints, and is validated for generalization across diverse EV models. The key contributions of this article are as follows:

\begin{table*}[t]
    \centering\caption{Relevant features according to different hysteresis models.}
    \renewcommand{\arraystretch}{1.2}
    
    \begin{tabular}{lcccccc}
        \specialrule{.1em}{.05em}{.05em} 
        \textbf{Model} & \textbf{Hysteresis Factor} & \textbf{Battery Current} & \textbf{Capacity} & \textbf{SoC} & \textbf{OCV Hysteresis Curves} & \textbf{Cell Voltage} \\
        \hline 
        Plett \cite{Plett2004ExtendedEstimation} & \checkmark & \checkmark &  &  &  &  \\
        \hline
        Roscher \cite{Roscher2011DynamicBatteries} &  & \checkmark & \checkmark &  &  &  \\
        \hline
        Xie \cite{Xie2016EstimatingCharacteristic} &  &  \checkmark &  &  & \checkmark &  \\
        \hline
        Ko \cite{Ko2021AStudy} &  & \checkmark & \checkmark & \checkmark &  &  \\
        \hline
        Xie \cite{Xie2023State-of-ChargeModel} & \checkmark & \checkmark & \checkmark & \checkmark &  &  \\
        \hline
        Plett \cite{Plett2004ExtendedIdentification} &  &  &  & \checkmark & \checkmark &  \\
        \hline
        Baronti \cite{Baronti2015Open-CircuitBatteries} &  &  &  & \checkmark & \checkmark &  \\
        \hline
        Dong \cite{Dong2016OnlineMethod} &  &  &  & \checkmark  & \checkmark &  \\
        \hline
        Kim \cite{Kim2012OCVCell} &  &  &  & \checkmark &  & \checkmark \\
        \hline
        Antony \cite{Antony2023ABatteries} &  & \checkmark & \checkmark &  &  &  \\
        \hline
        Zhou \cite{Zhou2022SOCBattery} &  &  &  & \checkmark & \checkmark & \checkmark \\
        \hline
        Yu \cite{Yu2022StudyModeling} &  &  &  &  &  & \checkmark \\
        \hline
        Gao \cite{Gao2023DeterminationBatteries} &  &  &  &  & \checkmark & \checkmark \\
        \hline
        Li \cite{Li2023ExploringBatteries} &  &  &  &  \checkmark & & \checkmark \\
        \hline
        Xu \cite{Xu2020ImprovingTechnique} &  &  \checkmark &  &  &  & \checkmark \\
        \specialrule{.1em}{.05em}{.05em} 
    \end{tabular}
    
    \label{tab:hysteresis_models}
\end{table*}

\begin{itemize}
    \item We propose a data harmonization framework to systematically process heterogeneous driving cycles and provide a valuable analysis of fleet data, yielding insights into real-world battery dynamics.
    
    \item We formulate hysteresis factor prediction in silicon–graphite anode-based batteries as a probabilistic modeling task, systematically benchmarking statistical and deep learning approaches under computational constraints, thereby advancing methods for uncertainty-aware battery state estimation.
    
    \item We provide a systematic evaluation of multiple transfer learning strategies (retraining, zero-shot prediction, fine-tuning, and joint training) across vehicle models with different cell chemistries and electrical specifications, deriving insights into the principles of model generalization for battery hysteresis prediction.
\end{itemize}

The paper is structured as follows:~\cref{harmonization} introduces the proposed data harmonization framework for heterogeneous driving cycles. Section \ref{model} presents the machine learning methods. The evaluation criteria are detailed in~\cref{metrics}. The experimental setup, along with the results for probabilistic hysteresis factor prediction, is discussed in~\cref{experiments}. Section \ref{conclusion} summarizes the findings and provides concluding remarks. Finally, \cref{limitation_future-work} describes limitations and future work.

\section{Data Harmonization Framework for Heterogeneous 
 Driving Cycles}
 \label{harmonization}
Let us consider driving cycles data consist of multiple time-series features, characterized by 
varying feature sets,  mixture of active and inactive periods, 
different sampling rates, diverse temporal lengths, 
and non-uniform units and scales, 
leading to a heterogeneous data structure. In detail, the heterogeneous driving cycles are formulated as a feature matrix \( X = \{X^{(i)} \in \mathbb{R}^{T_i \times F_i} \}_{i=1}^D \), where \( D \) is the number of driving cycles, \( T_i \) is the number of time steps (or sequence length) in the sample \( i \), and \( F_i \) is the number of features. Note that both \( T_i \) and \( F_i \) may vary across samples, reflecting differences in duration and available features.
We propose an approach to harmonize the heterogeneous driving cycles, illustrated in Fig. \ref{hdc} \textbf{a}.

\subsection{Data Filtering}
Various features are available, such as battery current, cell voltage, and SoC. 
Table~\ref{tab:hysteresis_models} overviews the features used in different hysteresis models from the literature. In alignment with prior work and engineering considerations, features influenced by the hysteresis factor and battery current, such as SoC and capacity, are excluded to minimize algorithmic bias. Additionally, OCV hysteresis curves are excluded, as these curves are only used as a precursor to calculate voltage changes, and this work predicts the hysteresis factor directly. Moreover, the cell temperature is included, as some models from the literature indirectly considered the temperature by performing parameter identification under different temperature conditions. 
Therefore, the input features considered in this work are battery current, cell voltage, and cell temperature.

The analysis of these features extracted from two vehicle models is shown in Fig.~\ref{hdc} \textbf{b}--\textbf{e}.
Note that the battery current is used as a proxy for the cell current. 
Furthermore, if an identified feature, such as cell temperature or cell voltage, is missing, the corresponding data sample is excluded.
Therefore, the feature matrix is updated to \( X' = \{X'^{(i)} \in \mathbb{R}^{T_i \times F} \}_{i=1}^{D'} \), where \( F \) becomes constant between samples, and \( D' < D \) as only partial samples contain all the needed features.

\subsection{Data Segmentation}

\begin{figure}[ht]
    \centering
    \includegraphics[width=0.48\textwidth]{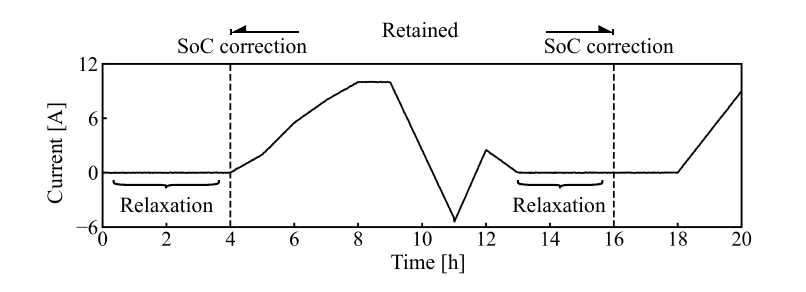}
    \caption{Example illustration of segmentation. A valid segment starts at the 4th hour, where a relaxation phase is followed by a SoC correction. The segment between hours 16 and 20 is not retained, as it lacks a closing SoC correction.}
    \label{fig:relaxation}
\end{figure}

Driving cycles often contain inactive periods, such as vehicle standby or battery relaxation phases. We apply a data segmentation strategy to identify and retain relevant data segments, illustrated in Fig. \ref{fig:relaxation}.  
A valid segment starts after a SoC correction if a relaxation phase occurs, during which the battery rests without charge or discharge, allowing the voltage to stabilize toward its open-circuit value. 
The segment is retained until the next SoC correction occurs. Any segment where the battery current remains zero throughout or the duration is shorter than 10 seconds is discarded. Each segmented data instance is treated as a separate sample. Therefore, the number of samples after segmentation is higher than \( D'\), as one driving cycle may contain several segments. This segmentation results in a new feature matrix \( X'' = \{X''^{(i)} \in \mathbb{R}^{T'_i \times F} \}_{i=1}^{N} \), where \( N \) is the number of samples after segmentation.

\subsection{Data Transformation}
\label{sec:datatransformation}

Some driving cycles may originally be sampled at higher or lower rates due to differences in data logging equipment, vehicle models, or testing protocols, causing temporal inconsistency.
To ensure uniformity and comparability across the dataset, all features are resampled to a fixed frequency of 10\, Hz (10 time steps per second).

\begin{figure}[h]
    \centering
    \includegraphics[width=0.48\textwidth]{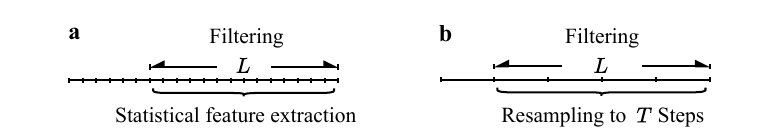}
    \caption{Illustration of two transformation approaches. 
    \textbf{a} Transformation for statistical learning models. 
    \textbf{b} Transformation for deep learning models.}
    \label{fig:transform_trunc}
\end{figure}

In addition, the segmented data described in the previous subsection consists of samples with varying sequence lengths. 
The variability in sequence length poses challenges for machine learning models, as different models expect different input formats. We introduce two data transformation approaches 
depending on the requirements of the target model, shown in Fig. \ref{fig:transform_trunc}.

\subsubsection{For Statistical Learning Models}
Statistical learning models require a two-dimensional input matrix. 
First, a \textit{temporal filter} is applied on \( X''^{(i)} \) to keep only the last \( L  \) time steps from each sample. This step results in a new feature matrix \( X''' \in \mathbb{R}^{N \times L \times F} \), where all sequences are truncated to a consistent temporal length across samples.
The temporal filter helps to study the optimal look-back window size given constrained computational power. 
Then, the statistical feature matrix \( X_s \in \mathbb{R}^{N \times F_s} \) is produced by computing \textit{statistical features}, such as mean, min, max, summarized in Table \ref{tab:time_series_attributes}, from each feature along the temporal dimension.

\begin{table}[ht]
\centering
\renewcommand{\arraystretch}{1.5}
\caption{Statistical features and their mathematical formulation.}
\begin{tabular}{ll}
\specialrule{.1em}{.05em}{.05em}
\textbf{Name} & \textbf{Mathematical definition} \\
\hline
Minimum value & $\displaystyle \min^{L}_{\ell=0} x_\ell$ \\ \hline
Maximum value & $\displaystyle \max^{L}_{\ell=0} x_\ell$ \\ \hline
Absolute minimum value & $\displaystyle \min^{L}_{\ell=0} |x_\ell|$ \\ \hline
Absolute maximum value & $\displaystyle \max^{L}_{\ell=0} |x_\ell|$ \\ \hline
Sum of absolute changes & $\displaystyle \sum^{L-1}_{\ell=0} |x_{\ell+1}-x_\ell|$ \\ \hline
Mean of changes & $\displaystyle\frac{1}{L}\sum^{L-1}_{\ell=0}(x_{\ell+1}-x_\ell)$ \\ \hline
Mean of absolute changes & $\displaystyle\frac{1}{L}\sum^{L-1}_{\ell=0}|x_{\ell+1}-x_\ell|$ \\ \hline
Absolute energy & $\displaystyle \sum^{L}_{\ell=0} x_\ell^2$ \\ \hline
Sum of values & $\displaystyle \sum^{L}_{\ell=0} x_\ell$ \\ \hline
Mean & $\displaystyle\frac{1}{L}\sum^{L}_{\ell=0} x_\ell$ \\ \hline
Complexity & $\displaystyle \sqrt{\sum^{L-1}_{\ell=0}(x_{\ell+1}-x_\ell)}$ \\ \hline
Sum of positive values & $\displaystyle\sum^{L}_{\ell=0}x_\ell, \text{ where } x_\ell>0$ \\ \hline
Sum of negative values & $\displaystyle\sum^{L}_{\ell=0}x_\ell, \text{ where } x_\ell<0$ \\ \hline
Mean of positive values & $\displaystyle\frac{1}{L}\sum^{L}_{\ell=0}x_\ell, \text{ where } x_\ell>0$ \\ \hline
Mean of negative values & $\displaystyle\frac{1}{L}\sum^{L}_{\ell=0}x_\ell, \text{ where } x_\ell<0$ \\ \hline
Number of zero elements & $\displaystyle \sum^{L}_{\ell=0} \mathbf{I}(x_\ell=0)$ \\ \hline
Root mean square & $\displaystyle \sqrt{\frac{1}{L}\sum^{L}_{\ell=0} x_\ell^2}$ \\ \hline
First value & $x_0$ \\ \hline
Last value & $x_L$ \\
\specialrule{.1em}{.05em}{.05em}
\end{tabular}
\label{tab:time_series_attributes}
\end{table}

\subsubsection{For Deep Learning Models}
\label{truncateDL} Deep learning models for time series often require a three-dimensional tensor. 
The same temporal filter is applied to \( X'' \) to produce \( X''' \). The statistical feature extraction is unnecessary since deep learning models inherently learn non-linear representations. However, under limited computational power, it is essential to reduce the resolution of the data. To this end, the truncated feature matrix \( X''' \) is resampled at a fixed number of evenly spaced time steps \( T \), resulting in a uniform 3D input tensor \( X_{d} \in \mathbb{R}^{N \times T \times F} \). The hyperparameter \( T \) controls the granularity of the time series representation: smaller values of \( T \) capture broader temporal patterns, while larger values retain finer dynamics. All combinations of \( L \) and \( T \) will be evaluated empirically to identify the optimal configuration for model performance.

\subsection{Data Scaling}
\label{scaling}

Features with different units and scales may introduce unintended biases, with models implicitly assigning greater importance to larger-magnitude inputs. Scaling also improves numerical stability and accelerates convergence during training \cite{Ioffe2015BatchShift}.

To ensure uniformity, all features are standardized to consistent units—for example, cell voltage is converted from millivolts (mV) to volts (V). Similar unit adjustments are applied to battery current, eliminating mismatches and enhancing comparability across samples.

Subsequently, we normalize the feature values to the range \([0, 1]\) using min–max scaling:

\subsubsection{For Statistical Learning Models}
We apply the standard \texttt{MinMaxScaler} implementation from \texttt{scikit-learn} to obtain the scaled feature matrix \( \tilde{X}_s \), ensuring consistent and reproducible feature scaling.

\subsubsection{For Deep Learning Models} 
We implement a customized min-max scaler. Given the input tensor \( X_d \in \mathbb{R}^{N \times T \times F} \), the scaled data \( \tilde{X}_d \) are calculated as:  

\begin{equation}
\tilde{X}_d = \frac{X_d - X_d^{\min}}{X_d^{\max} - X_d^{\min}},
\end{equation}

where \( X_d^{\min}, X_d^{\max} \in \mathbb{R}^{1 \times 1 \times F} \) represent the minimum and maximum values of each feature. This ensures that the feature matrix is scaled within \([0,1]\).

\section{Machine Learning Methods}
\label{model}
To enable uncertainty-aware predictions, all models in this work are trained to predict multiple target quantiles instead of a single point estimate.
Due to limited hardware resources, we apply three machine learning models for comparison, categorizing them into two types. The first type, statistical learning models, includes Linear Quantile Regression (LQR) and Quantile XGBoost (QXGB). These models operate on two-dimensional statistical feature inputs and are combined with dimensionality reduction techniques, namely Principal Component Analysis (PCA) and F-regression, to either compress or select features prior to modeling. The second type is a deep learning model, the Quantile Gated Recurrent Unit (QGRU), which processes three-dimensional sequential inputs without the need for statistical feature extraction. The objective is to identify the optimal model configuration while evaluating its computational requirements under constrained hardware resources, as typically encountered in vehicle applications.

\subsection{Principal Component Analysis (PCA)}
\label{sec:pca}

Principal Component Analysis (PCA) is used to reduce the dimensionality of the scaled statistical feature matrix \( \tilde{X}_s \in \mathbb{R}^{N \times F_s} \) by projecting it onto a lower-dimensional orthogonal subspace that captures the directions of maximum variance. 
Let \( W_{\mathrm{PCA}} \in \mathbb{R}^{F_s \times K} \) be the matrix of the top \( K \) eigenvectors of the empirical covariance matrix of \( \tilde{X}_s \), where \( K < F_s \). 
The choice of \( K \) is determined based on explained variance.

After dimensionality reduction, the compressed features are given by:
\begin{equation}
\tilde{X}_{\mathrm{PCA}} = \tilde{X}_s W_{\mathrm{PCA}},
\end{equation}
where \( \tilde{X}_{\mathrm{PCA}} \in \mathbb{R}^{N \times K} \) is then used as the input to statistical models such as LQR or QXGB.

This method decorrelates the input features and often improves the robustness and generalization of downstream models by eliminating noise and redundant information.
PCA also enables more efficient storage and faster computation, which is particularly beneficial in memory-constrained environments such as in-vehicle systems \cite{knodler2023potential}.

\subsection{F-Regression}
\label{sec:fregression}

F-regression evaluates the linear dependency between each feature and the target using the F-statistic. 
For each feature \( \tilde{x}_j \in \mathbb{R}^N \) (i.e., the \( j \)-th column of \( \tilde{X}_s \)), we perform a linear regression against the target \( y \in \mathbb{R}^N \) and compute the F-statistic \( \mathcal{F}_j \) and p-value \( p_j \):

\begin{equation}
\mathcal{F}_j = \frac{\mathrm{Var}(\mathbb{E}[y \mid \tilde{x}_j])}{\mathbb{E}[\mathrm{Var}(y \mid \tilde{x}_j)]}, \quad p_j = \mathbb{P}(\mathcal{F} > \mathcal{F}_j)
\end{equation}

where a smaller p-value \( p_j \) indicates stronger statistical relevance to the target. The top \( K \) features with the lowest p-values are selected to form the reduced matrix \( \tilde{X}_{\mathrm{F\text{-}reg}} \in \mathbb{R}^{N \times K} \), which is used as the input to LQR or QXGB.

This method allows for a simple yet effective feature selection process, prioritizing features that are most predictive under a linear assumption.
Unlike PCA, which produces linear combinations of all features, F-regression retains the original feature meanings, making the model more interpretable.
Furthermore, by discarding weakly correlated features, F-regression can help mitigate overfitting and reduce computational cost, particularly when the original feature set is large.

\subsection{Linear Quantile Regression (LQR)}
\label{sec:LQR}

LQR is chosen as a simple yet effective baseline due to its low computational complexity, fast training time, and interpretability \cite{koenker1978regression}. 
Unlike ordinary least squares (OLS), which estimates the conditional mean of the target variable, LQR directly estimates conditional quantiles, making it well-suited for probabilistic prediction tasks under high-variance input data.
It models the \(\tau\)-th conditional quantile of the target as a linear function of the input features:

\begin{equation}
\hat{y}_{\tau, i} = \tilde{X}_\mathrm{red}^T \beta_{\tau},
\end{equation}

where \( \tilde{X}_\mathrm{red} \in \mathbb{R}^{N \times K} \) represents the reduced feature matrix obtained from either PCA (\( \tilde{X}_{\mathrm{PCA}} \)) or F-regression (\( \tilde{X}_{\mathrm{F\text{-}reg}} \)), and \( \beta_{\tau} \in \mathbb{R}^K \) is the parameter vector estimated for quantile \( \tau \).

The model parameters are learned by minimizing the quantile loss \( L_\tau(y_i, \hat{y}_{\tau, i}) \) described in Section~\ref{sec:AQL}, which penalizes under- and over-predictions asymmetrically depending on the target quantile. 

Due to its linear structure, LQR can be trained efficiently even on large datasets and provides easily interpretable coefficients that reflect the marginal effect of each feature on different parts of the target distribution.
However, its modeling capacity is limited by the assumption of linearity, making it less flexible than more complex models such as tree-based methods or neural networks.

\subsection{Quantile XGBoost (QXGB)}

\(\mathrm{QXGB}\) is selected due to its strong performance in capturing nonlinear relationships by leveraging gradient-boosted decision trees \cite{chen2016xgboost}. The predicted quantile is given by:
\begin{equation}
\hat{y}_{\tau, i} = f_{\tau}(\tilde{X}_\mathrm{red}),
\end{equation}
where \( \tilde{X}_\mathrm{red} \in \mathbb{R}^{N \times K} \) is the reduced feature matrix produced by PCA or F-regression, and \( f_{\tau} \) denotes an ensemble of boosted decision trees trained for quantile \( \tau \). Specifically, the model consists of \( M \) additive trees:
\begin{equation}
f_{\tau}(\tilde{X}_\mathrm{red}) = \sum_{m=1}^{M} h_m(\tilde{X}_\mathrm{red}),
\end{equation}
where each base learner \( h_m \) is trained iteratively to minimize the quantile loss \( L_\tau(y_i, \hat{y}_{\tau, i}) \) as defined in Section~\ref{sec:AQL}.

\subsection{Quantile Gated Recurrent Unit (QGRU)}

Recurrent neural networks (RNNs) are well-suited for time series data. Although both QLSTM and QGRU can model sequential dependencies, QGRU has been shown to be more parameter-efficient \cite{chung2014empirical}, making it a better choice given our hardware constraints.  

The QGRU model processes scaled sequential inputs \( \tilde{X}_d \in \mathbb{R}^{N \times T \times F} \). The update equations for the QGRU are:
\begin{align}
z_t &= \sigma(W_z \tilde{X}_t + U_z h_{t-1} + b_z), \\
r_t &= \sigma(W_r \tilde{X}_t + U_r h_{t-1} + b_r), \\
\tilde{h}_t &= \tanh(W_h \tilde{X}_t + U_h (r_t \odot h_{t-1}) + b_h), \\
h_t &= (1 - z_t) \odot h_{t-1} + z_t \odot \tilde{h}_t,
\end{align}
where \( \tilde{X}_t \in \mathbb{R}^{N \times F} \) represents the input at time step \( t \), and \( h_t \in \mathbb{R}^{N \times H} \) denotes the hidden state. The learnable parameters include the weight matrices \( W_z, W_r, W_h \in \mathbb{R}^{F \times H} \) and \( U_z, U_r, U_h \in \mathbb{R}^{H \times H} \), and the bias terms \( b_z, b_r, b_h \in \mathbb{R}^{H} \). The activation functions are the sigmoid function \( \sigma(\cdot) \) for the update gate \( z_t \) and reset gate \( r_t \), and the hyperbolic tangent function \( \tanh(\cdot) \) for computing the candidate hidden state \( \tilde{h}_t \). The operator \( \odot \) denotes element-wise multiplication.

The predicted quantile is obtained as:
\begin{equation}
\hat{y}_{\tau, i} = g_{\tau}(h_T),
\end{equation}
where \( g_{\tau} \) is a fully connected layer mapping the final hidden state \( h_T \) to the quantile output.

The QGRU model is trained by minimizing the same quantile loss \( L_\tau(y_i, \hat{y}_{\tau, i}) \) from Section~\ref{sec:AQL}.

\section{Metrics}
\label{metrics}
We consider three key metrics: Average Quantile Loss (AQL), Random Access Memory (RAM), and Read-Only Memory (ROM) to assess the model’s prediction performance and computational efficiency. 

\subsection{Average Quantile Loss (AQL)}
\label{sec:AQL}
AQL is utilized to estimate the conditional quantiles of the target distribution \cite{koenker2005quantile}. Given a quantile level \( \tau \in (0, 1) \), the quantile loss function \( L_\tau \) is defined as:
\begin{equation}
    L_\tau(y_i, \hat{y}_{\tau, i}) = 
    \begin{cases} 
      \tau \cdot (y_i - \hat{y}_i), & \text{if } y_i \geq \hat{y}_i, \\
      (1 - \tau) \cdot (\hat{y}_i - y_i), & \text{if } y_i < \hat{y}_i,
    \end{cases}
\end{equation}
where \( y_i \) denotes the true value, and \( \hat{y}_i \) represents the predicted quantile for the \( i \)-th sample. This loss function asymmetrically penalizes errors: when predicting upper quantiles (\( \tau > 0.5 \)), under-predictions receive a higher penalty, whereas for lower quantiles (\( \tau < 0.5 \)), over-predictions are penalized more. 

The AQL is calculated as the mean quantile loss over all predicted quantiles and samples:
\begin{equation}
    \text{AQL} = \frac{1}{|\mathcal{Q}| N} \sum_{\tau \in \mathcal{Q}} \sum_{i=1}^N L_\tau(y_i, \hat{y}_{\tau, i}),
\end{equation}
where \( \mathcal{Q} \) is the set of quantile levels considered, and \( N \) denotes the total number of samples. A lower AQL value indicates improved overall quantile prediction performance.

\subsection{Random Access Memory (RAM)}
RAM is used to measure the real-time computational workload and data handling efficiency of machine learning models, measured in megabytes (MB). As a volatile memory, RAM temporarily stores intermediate computations, activations, and model states during training and inference. In the context of EVs, hardware resources are limited, making RAM consumption a critical factor for deploying machine learning models. An efficient model should minimize RAM usage while maintaining predictive performance, ensuring compatibility with on-board computing limitations.

\begin{table}[t]
\caption{Hyperparameter search range.}
\label{tab:dl_params}
\begin{center}
\begin{small}
\begin{tabular}{lll}
\toprule
\textbf{Model} & \textbf{Hyperparameter} & \textbf{Range}  \\
\midrule

LQR & Reg. L1        & [0, 1]  \\
\midrule

QXGB & Learning Rate         & [0.005, 0.1] \\
 & Max. Depth             & [3, 5]           \\
 & Subsample             & [0.6, 1]          \\
 & Reg. Lambda            & [0.1, 10]        \\
 & Reg. Alpha             & [0.1, 3]         \\
 & Min. Child Weight      & [1, 4]  \\
 & Colsample by Tree     & [0.7, 0.9]  \\
 & Num. Boost Rounds      & [50, 1000]  \\
\midrule

QGRU & Learning Rate    & [0.0001, 0.01]  \\
 & Hidden Size      & [4, 64]  \\
 & Num. Layers       & [1, 2]  \\
 & Batch Size       & [8, 32]  \\
\bottomrule
\end{tabular}
\end{small}
\end{center}
\end{table}

\subsection{Read-Only Memory (ROM)}
ROM is used to measure the storage requirements of machine learning models, particularly for deployment in resource-limited environments, measured in MB. Unlike RAM, ROM is non-volatile, meaning it retains data even after power loss. It stores the trained model’s weights, biases, and static parameters required for inference. For the hysteresis factor prediction in batteries, minimizing ROM usage is essential to fit within the storage constraints of EV hardware. A compact model with low ROM requirements ensures efficient deployment, preserving space for other essential vehicle control and battery management functions. 


\begin{figure*}[t]
\begin{center}
\centerline{\includegraphics[width=1.0\linewidth]{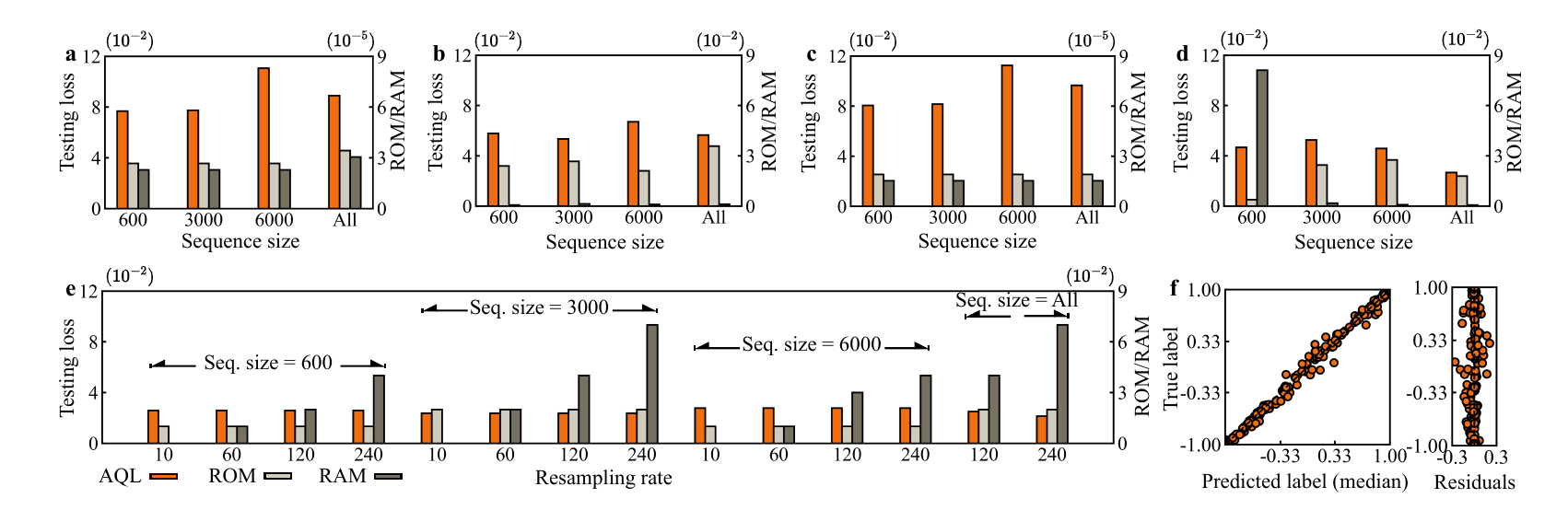}}
\caption{
Comparison of model performance and computational efficiency for different modeling strategies on sequence-level and subsequence-level prediction tasks.
\textbf{a} PCA + LQR on sequence-level prediction. 
\textbf{b} PCA + QXGB on sequence-level prediction. 
\textbf{c} F-Reg + LQR on sequence-level prediction. 
\textbf{d} F-Reg + QXGB on sequence-level prediction. 
\textbf{e} QGRU on sequence-level prediction. Different resampling rates are explored for each sequence size. 
\textbf{f} Scatter plots of true label versus predicted (median) label (left) and residuals (right) for QGRU on subsequence-level prediction. Tight clustering along the diagonal and low residuals demonstrate the strong predictive performance of QGRU.
}
\label{exp_prob}
\end{center}
\end{figure*}

\begin{table*}[t]
\label{tab:metrictable}%
\begin{center}
\caption{Model performance comparison on testing data.}
\begin{small}
\begin{tabular}{lllllll}
\multicolumn{7}{l}{Sequence-Level Prediction} \\
\toprule
Dim. Red. & Model & Seq. Size & Resamp. Rate & AQL $\downarrow$ & ROM $\downarrow$ & RAM $\downarrow$ \\
\midrule
PCA & LQR & 600 & -- &  \(7.67\times 10^{-2}\) & \(2.67\times 10^{-5}\) & \(2.29\times 10^{-5}\) \\
& QXGB & 3000 & -- & \(5.34\times 10^{-2}\) & \(2.67\times 10^{-2}\) & \(1.35\times 10^{-3}\) \\
\midrule
F-Reg & LQR & 600 & -- & \(8.06\times 10^{-2}\) & \(\bm{1.91\times 10^{-5}}\) & \(\bm{1.53\times 10^{-5}}\) \\
& QXGB & All & -- & \(2.67\times 10^{-2}\) & \(1.79\times 10^{-2}\) & \(6.80\times 10^{-4}\) \\
\midrule
-- & QGRU & All & 240 & \(\bm{2.15\times 10^{-2}}\) & \(2.00\times 10^{-2}\) & \(7.00\times 10^{-2}\) \\
\bottomrule
 &  &  &  &  \\
\multicolumn{7}{l}{Subsequence-Level Prediction} \\
\toprule
Dim. Red. & Model & Subseq. Size & Resamp. Rate & AQL $\downarrow$ & ROM $\downarrow$ & RAM $\downarrow$ \\
\midrule
-- & QGRU & 6000 & 240 & \(\bm{2.68\times 10^{-3}}\) & \(\bm{1.00\times 10^{-2}}\) & \(\bm{4.00\times 10^{-2}}\) \\
 & QGRU$^{*}$ & 6000 & 240 & \(4.65\times 10^{-3}\) & \(1.00\times 10^{-2}\) & \(4.00\times 10^{-2}\) \\
\bottomrule

\end{tabular}
\end{small}
\end{center}
\vskip -0.1in
\end{table*}

\section{Probabilistic Hysteresis Factor Prediction}
\label{experiments}

The experiment is conducted in two stages: first, a \textit{sequence-level prediction} is performed to identify the optimal model and its configuration. Then, the identified model is applied in a \textit{subsequence-level prediction} setting. The target quantiles are set at 5\%, 50\%, and 95\%, respectively. The dataset is split into 80\%, 10\%, and 10\% for training, validation, and testing, respectively.
The hyperparameters are optimized based on an exhaustive random grid search with 5-fold cross-validation, employing scikit-learn's RandomizedSearchCV. The search space of hyperparameters is listed in Table \ref{tab:dl_params}, totalling more than 1,000 trials.

\subsection{Sequence-Level Prediction}

In this approach, each sample includes the full sequence. Each sequence is truncated to keep the last \( L \in \{600, 3000, 6000, All\} \) time steps, corresponding to \( S \in \{60, 300, 600, All\} \) seconds. 

Each truncated sequence is resampled at a fixed number of time steps \( T \in \{10, 60, 120, 240\} \) to ensure consistent input dimensions, as described in~\cref{truncateDL}.

The performance comparison is summarized in Table \ref{tab:metrictable}, where the best results are highlighted in \textbf{bold}.
Note that the table only reports the best configuration for each model, i.e., the combination of sequence size and resampling rate that yields the lowest AQL. 
Further details on the full set of configurations and their performance can be found in Fig. \ref{exp_prob}.

Among statistical learning models, QXGB with F-Reg achieves a strong trade-off, reaching an AQL of \(2.67 \times 10^{-2}\) with ROM and RAM consumption of \(1.79 \times 10^{-2}\) MB and \(6.80 \times 10^{-4}\) MB, respectively, making it a competitive choice for resource-constrained scenarios.
In contrast, LQR models show poorer predictive performance, with an AQL value of \(7.67 \times 10^{-2}\). Moreover, increasing the sequence size does not lead to a consistent decrease in testing loss, indicating that the LQR is unable to fully leverage longer input sequences. However, LQR can still be viable for extremely low-resource applications given its minimal ROM and RAM usage.
Notably, QGRU achieves the lowest AQL of \(2.15 \times 10^{-2}\) while maintaining reasonable ROM and RAM usage. This demonstrates the strong representational capacity of deep learning models compared to traditional statistical approaches. It is seen that the optimal resampling rate of 240 yields the lowest AQL for QGRU. Increasing the resampling rate could potentially further reduce AQL; however, this may result in significantly higher memory usage, which could be prohibitive in resource-constrained environments.

Overall, the QGRU stands out as the most accurate model with a tolerable computational cost, whereas QXGB demonstrates a favorable trade-off between predictive performance and memory usage, making it suitable for moderately constrained environments. LQR models are more appropriate for extremely limited applications, where computational efficiency is prioritized over predictive performance.

\subsection{Subsequence-Level Prediction}
\label{sec:subseq_pred}
In practical deployments, hysteresis factors are predicted continuously over time. To mimic this setting, each sample is firstly divided into fixed-length subsequences, rather than applying a temporal filter to truncate the sequences, as introduced in~\cref{truncateDL}. Building on the previous subsection, where QGRU demonstrated the best AQL loss, we further evaluate QGRU for subsequence-level prediction. In detail, each subsequence spans a fixed length of \( L \in \{100, 600, 3000, 6000\} \) time steps from each sample, corresponding to \( S \in \{10, 60, 300, 600\} \) seconds. Then, the subsequences are similarly resampled at a fixed number \( T \in \{10, 60, 120, 240\} \). Each subsequence inherits the same label as in the sequence-level prediction setting, where a single label is assigned to the entire sequence.

Results in Table \ref{tab:metrictable} show that the best configuration uses subsequences of 6000 time steps with a resampling rate of 240, which satisfies the hardware requirements of our setup—ROM under \(1 \times 10^{-2}\) MB, and RAM under \(4 \times 10^{-2}\) MB. This configuration offers a balance between predictive performance and deployment feasibility. It is worth noting that further improvements may be achieved by increasing the time steps or the resampling rate. 

We further evaluate QGRU under two configurations: non-autoregressive and autoregressive. In the non-autoregressive setting, each subsequence is treated independently. In the autoregressive setting, temporal dependencies are introduced by feeding the previous subsequence’s prediction as an additional input feature to the next. 

Results in Table \ref{tab:metrictable} indicate that the QGRU$^{*}$ (autoregressive) yields higher test loss than QGRU (non-autoregressive). This suggests that the accumulation of prediction errors over subsequences degrades performance 
and is not sufficiently compensated by implicitly learned temporal covariates. We regard high input variance data in combination with the quantile task setting as the possible explanation.

\section{Generalization Across Vehicle Models}

This section evaluates the generalization capability of the best model (QGRU) under unseen vehicle models. Specifically, the vehicle model~B is introduced, as visualized in Fig. \ref{hdc}.
Both models correspond to fully EVs; however, they differ in terms of cell chemistry and electrical specifications \footnote{Due to commercial confidentiality, specifics regarding the cell chemistry and electric specifications cannot be disclosed.}. We consider the following evaluation strategies to systematically study cross-model performance.

\begin{enumerate}
    \item B/B (retraining): Train on dataset~B, and test on dataset~B, evaluating in-distribution generalization.
    
    \item A/B (zero-shot prediction): Train on dataset~A, and test on dataset~B, evaluating out-of-distribution generalization.

    \item A→B/B (fine-tuning): Train on dataset~A,  fine-tune on dataset~B, and test on dataset~B, assessing generalization performance via fine-tuning.
    
    \item A+B/B (joint training): Train on the combined dataset (A + B), and test on dataset~B, assessing joint distribution generalization.

\end{enumerate}

Moreover, we compare the impact of the data scaler across different evaluation strategies. As the cell chemistries in vehicle models~A and~B differ, their corresponding voltage ranges also vary, resulting in different scaling behaviors when using a min-max scaler. For retraining (B/B) and zero-shot prediction (A/B) scenarios, we evaluate the effect of applying the old scaler fitted on training data from dataset~A versus refitting a new scaler using training data from dataset~B. For fine-tuning (A$\rightarrow$B/B) or joint training (A+B/B) scenarios, the new scaler is refitted using the combined training data, and thus reflects the minimum and maximum values from both datasets. Note that in all cases, model hyperparameters are optimized using the respective validation sets.

\begin{table}[t]
\label{tab:scaler_comparison}
\begin{center}
\caption{Comparison of testing loss under different training strategies and scalers.}
\begin{small}
\begin{tabular}{llll}
\toprule
Strategy & Scaler & AQL $\downarrow$ & 
\textcolor{black}{Change $\downarrow$} \\
\midrule
B/B & Old  & \(7.86 \times 10^{-3}\) & \textcolor{black}{+0.0\%} \\
    & New  & \(8.38 \times 10^{-3}\) & \textcolor{black}{+6.2\%} \\
\midrule
A/B & Old  & \(4.06 \times 10^{-2}\) & \textcolor{black}{+415.9\%} \\
    & New  & \(6.49 \times 10^{-2}\) & \textcolor{black}{+726.0\%} \\
\midrule
A$\rightarrow$B/B & Old & \(8.62 \times 10^{-3}\) & \textcolor{black}{+9.7\%} \\
                  & New & \(8.23 \times 10^{-3}\) & \textcolor{black}{+4.7\%} \\
\midrule
A+B/B & Old & \(\bm{7.60 \times 10^{-3}}\) & \textcolor{black}{\textminus3.3\%} \\
      & New & \(8.82 \times 10^{-3}\) & \textcolor{black}{+12.2\%} \\
\bottomrule
\end{tabular}
\end{small}
\end{center}
\vskip -0.1in
\end{table}

The Table \ref{tab:scaler_comparison} presents the resulting testing losses under each scenario. Training exclusively on dataset~A and testing on dataset~B results in the highest testing loss, with 415.9\% higher loss than training solely on dataset~B with the old scaler. 
Fine-tuning on B or combining A and B for training offers improvements compared to solely training on dataset~A. 

Additionally, using a new scaler fitted on dataset~B often underperforms compared to the old scaler solely fitted on dataset~A. This trend can be attributed to the fact that dataset~A is significantly larger, and the derived statistics provide a more representative feature scaling. Poor prediction performance may also result from the differences in cell chemistry between vehicle project~B and the training data from project~A. 
If a new scaler is fitted using dataset~B, but the model parameters remain unchanged, the transformed input distribution becomes misaligned with the model's learned representation.
This distribution shift can hinder the model's ability to interpret the inputs correctly, leading to degraded performance.
In the fine-tuning scenario (A$\rightarrow$B/B), where the model is not trained from scratch but adapted from pretrained weights on dataset~A, the new scaler can offer modest improvements. This suggests that when the model parameters are partially aligned with dataset~B, adjusting the feature scaling to match the target distribution becomes beneficial.

Moverover, the best loss solely with dataset~B (\(8.38 \times 10^{-3}\)) is significantly higher than the loss achieved for dataset~A (\(2.68 \times 10^{-3}\)). The discrepancy could be attributed to the availability of data, as less than 70\% of the data for vehicle model~A is available for vehicle model~B, as mentioned in Fig. \ref{hdc}. Therefore, if more data for vehicle model B are available, the predictive performance would also improve.

\section{Conclusion}
\label{conclusion}

This work addressed the challenge of modeling voltage hysteresis in silicon-graphite anode-based batteries. 
We introduced a data harmonization framework to process heterogeneous driving cycles under varying operating conditions and developed a probabilistic approach for hysteresis factor prediction. Unlike prior methods that often overlook predictive uncertainty and computational efficiency, our method explicitly accounts for uncertainty and computational constraints, making it well-suited for deployment in real-world EV applications.

Experimental results demonstrate that QGRU is the most suitable model when predictive performance is the primary concern. QXGB with F-Reg is recommended as a practical trade-off between accuracy and computation. In contrast, LQR offers minimal computational requirements but performs significantly worse in terms of AQL, making it suitable only for extremely constrained environments.
The generalization assessment demonstrates that zero-shot prediction performs significantly worse than transfer learning strategies and is therefore not recommended in practice, particularly when cell chemistries and electrical specifications differ.

\section{Limitations and Future Work}
\label{limitation_future-work}

A key limitation of this work lies in its dependency on the labeling algorithm used to construct learning targets, influencing the predictive performance of QGRU. 
Moreover, although QGRU achieves the best predictive performance, its computational cost is higher compared to simpler models, which may limit applicability under highly constrained hardware conditions. 
Furthermore, the common issue of data imbalance affects the performance of transfer learning, as models trained on skewed distributions may not generalize well to new vehicle chemistries. 
Future work should focus on improving labeling algorithms within BMS and on constructing chemistry-specific, cell-level datasets to enable a more rigorous evaluation of labeling quality. 
In addition, exploring advanced yet computationally efficient deep learning models is worthwhile to further improve predictive accuracy while reducing computational demand. Lastly, as hardware capabilities in electric vehicles continue to improve, it will become feasible and beneficial to increase both the (sub)sequence length and the resampling rate, thereby enhancing model expressiveness and accuracy while still meeting deployment requirements.

{\appendix
\subsection{Package Version}
The specific Python libraries are listed in Table~\ref{tab:lib_versions}.

\begin{table}[ht]
\label{tab:lib_versions}
\begin{center}
\caption{Key Python libraries and their respective version.}
\begin{small}
\begin{sc}
\begin{tabular}{ll}
\toprule
\textbf{Library} & \textbf{Version} \\
\midrule
\texttt{scikit-learn}  & 1.4.2 \\
\texttt{tsfresh}       & 0.20.2 \\
\texttt{xgboost}       & 2.0.3 \\
\texttt{torch}         & 2.3.1+cu121 \\
\texttt{ray}           & 2.34.0 \\
\bottomrule
\end{tabular}
\end{sc}
\end{small}
\end{center}
\vskip -0.1in
\end{table}

\subsection{Hardware Description}
 All experiments are conducted on a 64-bit Windows workstation equipped with an Intel\textsuperscript{\textregistered} Core\texttrademark{} i9-10920X CPU operating at a base frequency of 2.5\,GHz. The machine has 128\,GB of DIMM RAM and two NVIDIA RTX A5000 GPUs, each with 87.55\,GB of GPU memory.

\subsection{Confidentiality of Labeling Algorithm and Data}
In the labeling algorithm developed by the Porsche Engineering Group, a parameter is used that must be calibrated for each vehicle project, making the process time-consuming. This algorithm is highly commercially confidential; therefore, we are unable to disclose further technical details. 

The fleet data used in this work are subject to commercial confidentiality in the competitive automotive industry. To provide maximum transparency, we have established a \href{https://runyao-yu.github.io/Porsche_Hysteresis_Factor_Prediction/}{project website}. If the dataset becomes commercially non-sensitive in the future, we will update the website to release more detailed information.
}

\printbibliography[heading=bibintoc,title=References]

\vfill

\end{document}